\documentclass{article}
\usepackage{spconf,amsmath,graphicx}
\usepackage{amsfonts,amssymb,array}
\usepackage{latexsym}
\usepackage{CJK}
\usepackage{cite}
\usepackage{bm}
\usepackage{soul}

\usepackage{epstopdf}
\usepackage{epsfig}
\usepackage{subfigure}

\usepackage{mathrsfs}	
\usepackage{algorithm} 
\usepackage{algorithmic} 
\usepackage{booktabs}

\DeclareMathOperator*{\argmin}{argmin}
\newcommand{\nn}{\nonumber}
\title{Model change detection with application to machine learning}
\name{Yuheng Bu$^{\star }$   \qquad Jiaxun Lu $^{\dagger}$ \qquad  Venugopal V. Veeravalli $^{\star}$
\thanks{The work of Y. Bu and V. V. Veeravalli was supported by the Army Research
Laboratory under Cooperative Agreement W911NF-17-2-0196 through the University of Illinois at Urbana-Champaign. }}
\address{$^{\star}$ University of Illinois at Urbana-Champaign \qquad  $^{\dagger}$ Tsinghua University\\ Email: \small{ \texttt{bu3@illinois.edu,  lujx14@mails.tsinghua.edu.cn, vvv@illinois.edu}}}

\newtheorem{thm}{Theorem}
\newtheorem{assump}{Assumption}
\begin{document}
%
\maketitle
\begin{abstract}
Model change detection is studied, in which there are two sets of samples that are independently and identically distributed (i.i.d.) according to a pre-change probabilistic model with parameter $\theta$, and a post-change  model with parameter $\theta'$, respectively. The goal is to detect whether the change in the model is significant, i.e., whether the difference between the pre-change parameter and the post-change parameter $\|\theta-\theta'\|_2$ is larger than a pre-determined threshold $\rho$. The problem is considered in a Neyman-Pearson setting, where the goal is to maximize the probability of detection under a false alarm constraint. Since the generalized likelihood ratio test (GLRT) is difficult to compute in this problem, we construct an empirical difference test (EDT), which approximates the GLRT and has low computational complexity. Moreover, we provide an approximation method to set the threshold of the EDT to meet the false alarm constraint. Experiments with linear regression and logistic regression are conducted to validate the proposed algorithms.
\end{abstract}
%

\begin{keywords}
Model change detection, generalized likelihood ratio test, Neyman-Pearson setting
\end{keywords}
\section{Introduction}
\label{sec:intro}
We study the model change detection problem, where two sets of samples are independently and identically distributed (i.i.d.) according to a pre-change probabilistic model with parameter $\theta$, and a post-change probabilistic model with parameter $\theta'$, respectively. The goal is to determine whether the change in the model is significant or not. We formulate the problem in a Neyman-Pearson setting, and adopt the $\ell_2$ distance between the parameters to measure the change between the models. More specifically, our goal is to construct a test to detect whether $\|\theta-\theta'\|_2$ is larger than a pre-determined threshold $\rho$, while satisfying a false alarm constraint.

This problem is motivated in part by the recent works on active and adaptive sequential learning \cite{wilson2018adaptive,wilson2016adaptive,bu2018active}, where the machine learning models learned in previous time-steps are used adaptively to improve the accuracy and data-efficiency in the next time-step. A key step in applying these adaptive sequential learning methods is the detection of an abrupt or large model change, since
adapting to the previous model if it is significantly different from the current one could deteriorate performance. A specific application in this context  is the detection of a shift of user preferences in personalized recommendation systems \cite{elahi2016survey,rubens2015active}. In addition, we believe that our model change detection formulation can be applied in transfer learning \cite{pan2010survey} to determine whether two machine learning tasks are transferable.



We note that our model change detection problem is different from the quickest change detection problem studied in \cite{geng2016online,zou2017quickest}. There a linear regression model changes at an unknown point in time, and the goal is to detect the change as soon as possible with streaming data. We are interested in detecting whether the change in the model is significant, given sets of samples from the pre- and post-change models.

A standard method for solving a composite hypothesis testing problem such as the model change detection problem under consideration is the generalized likelihood ratio test (GLRT). However, the maximum likelihood estimates of $\theta$ and $\theta'$ required in the GLRT are difficult to compute under the constraint $\|\theta-\theta'\|_2\le \rho$ in this case. Our first contribution is to propose an empirical difference test (EDT), which approximates the GLRT and has low computational complexity. Moreover, we provide an approximation method to set the threshold in the proposed EDT, which ensures a bound on the worst-case false alarm probability. We validate our results using experiments involving linear regression and logistic regression.



\section{Problem Model}
\label{Sec:ProblemModel}
Throughout this paper, we use lower case letters to denote scalars and vectors, and use upper case letters to denote random variables and matrices. We use $\lambda_{\max}(A)$ and $\lambda_{\min}(A)$ to denote the largest and the smallest eigenvalues of matrix $A$, respectively, and $\mathrm{Tr}(A)$ to denote the trace of a square matrix $A$.  All logarithms are the natural ones. 


We consider the  model change detection problem in the following setting. We are given two datasets $\mathcal{S} = \{\mathbf{z}_1, \cdots, \mathbf{z}_n\}$ and $\mathcal{S}' = \{\mathbf{z}'_1, \cdots, \mathbf{z}'_{n'}\}$ with samples $\mathbf{z}$ drawn from some instance space $\mathcal{Z}$. In addition, we are given a parameterized family of distribution models $\mathcal{M}=\{ p(\mathbf{z}|\theta), \theta\in \mathbb{R}^d\}$. We assume that there exist two unknown parameters $\theta, \theta' \in \mathbb{R}^d$, such that the datasets $\mathcal{S}$ and $\mathcal{S}'$ are independently generated from the following pre-change and post-change models, respectively,
\begin{equation}
 \mathbf{Z}_i \sim p(\mathbf{z}_i|\theta),\ \mathbf{z}_i \in \mathcal{S}, \quad \mbox{and} \quad
 \mathbf{Z}'_j \sim p(\mathbf{z}'_i|\theta'),\ \mathbf{z}'_j \in \mathcal{S'}.
\end{equation}

Our goal is to construct a computational efficient test to decide between the following two hypotheses:
\begin{equation}\label{Equ:LRMHypothesis}
\begin{split}
H_0:  (\theta, \theta') &\in \chi_0 \triangleq \{ (\theta, \theta')|\,\|\theta- \theta'\|_2 \le \rho \}, \\
H_1:  (\theta, \theta') &\in \chi_1 \triangleq \{ (\theta, \theta')|\,\|\theta- \theta'\|_2 > \rho \},
\end{split}
\end{equation}
where $\rho$ is a constant determined by the specific applications.

Let $\delta: \mathcal{Z}^{n}\times \mathcal{Z}^{n'} \to \{0,1\}$ denote the decision rule for the model change detection problem. Then the probabilities of false alarm and correct detection can be written as
\begin{align}
  {\rm{P}}_{\rm{F}}(\delta,\theta,\theta') &\triangleq {\rm{P}}_{(\theta,\theta')}\{\delta(S,S')=1\},\quad \forall(\theta, \theta') \in\chi_0,\\
  {\rm{P}}_{\rm{D}} (\delta,\theta,\theta')&\triangleq {\rm{P}}_{(\theta,\theta')}\{\delta(S,S')=1\},\quad \forall(\theta, \theta') \in\chi_1,
\end{align}
where ${\rm{P}}_{(\theta,\theta')}$ denotes the probability measure for the data conditioned on the model parameter $(\theta,\theta')$.

Note that in \eqref{Equ:LRMHypothesis}, both the null hypothesis and the alternative hypothesis are composite. We study the detection problem in the Neyman-Pearson setting:
\begin{equation}\label{Equ:NPRule}
\begin{split}
\max_{\delta}\,\,\, &{\rm{P}}_{\rm{D}}(\delta,\theta,\theta'), \quad\quad\quad \forall(\theta, \theta') \in\chi_1\\
\text{s.t.}\,\,\,& {\rm{P}}_{\rm{F}}(\delta,\theta,\theta')\leq\alpha, \quad \forall(\theta, \theta') \in\chi_0.
\end{split}
\end{equation}
As seen in \eqref{Equ:NPRule}, our goal is to construct a test that maximizes the detection probability for all $(\theta, \theta') \in \chi_1$, and satisfies the false alarm constraint for all $(\theta, \theta') \in\chi_0$. The solution to \eqref{Equ:NPRule} if it exists is said to be a uniformly most powerful (UMP) test.

Since $\mathbf{z}_i$ and $\mathbf{z}'_i$ are drawn i.i.d. from $p(\mathbf{z}_i|\theta)$ and $p(\mathbf{z}_i'|\theta')$, respectively, we can use
\begin{equation}
  L(\theta) \triangleq -\sum_{i=1}^n \log p(\mathbf{z}_i|\theta),\quad  L'(\theta) \triangleq -\sum_{i=1}^{n'}\log p(\mathbf{z}'_i|\theta)
\end{equation}
to denote the negative log-likelihood functions with the pre-change dataset $\mathcal{S}$ and post-change dataset $\mathcal{S}'$, respectively. Then, the maximum likelihood estimates (MLE) of $\theta$ and $\theta'$ can be written as
\begin{equation}
\hat{\theta}_{\rm{ML}} \triangleq \argmin L(\theta), \qquad \hat{\theta}'_{\rm{ML}} \triangleq \argmin L'(\theta).
\end{equation}

In addition, we denote the Hessian matrices of $L(\theta)$ and $L'(\theta)$ as $H(\theta)\triangleq \nabla^2_{\theta}{ L(\theta)}$, and  $H'(\theta)\triangleq \nabla^2_{\theta}{ L'(\theta)}$.

\section{Empirical Difference Test}
\label{Sec:EDT}
\subsection{Generalized Likelihood Ratio Test}
In general, a UMP solution to the composite hypothesis testing problem in \eqref{Equ:NPRule} may not exist, and may be difficult to find even if it exists. An alternative approach is to apply the GLRT. The generalized log-likelihood ratio (GLR) is given by
\begin{small}
\begin{align}
L_{\rm{G}}(\mathcal{S},\mathcal{S}') &\triangleq \log \frac{\max_{(\theta,\theta')\in\chi_1} \prod_{i=1}^n p(\mathbf{z}_i|\theta)\prod_{i=1}^{n'} p(\mathbf{z}'_i|\theta)}{\max_{(\theta,\theta')\in\chi_0} \prod_{i=1}^n p(\mathbf{z}_i|\theta)\prod_{i=1}^{n'} p(\mathbf{z}'_i|\theta)}.
\end{align}
\end{small}

If $L_{\rm{G}}(\mathcal{S},\mathcal{S}')$ does not have point masses under either $H_0$ or $H_1$, the GLRT has the following structure
\begin{equation}\label{Equ:GLTest}
\delta_{\rm{GL}}(\mathcal{S},\mathcal{S}') =
\begin{cases}
1,\quad \text{if} \,\, L_{\rm{G}}(\mathcal{S},\mathcal{S}') \geq \tau\\
0,\quad \text{if} \,\, L_{\rm{G}}(\mathcal{S},\mathcal{S}') < \tau,
\end{cases}
\end{equation}
where $\tau$ is the threshold for the GLR statistics determined by the false alarm constraint $\alpha$.

For the conciseness, we define
\begin{equation}\label{Equ:GLopt}
\begin{split}
  (\hat{\theta}_1, \hat{\theta}'_1) \triangleq \argmin_{(\theta,\theta')\in\chi_1} L(\theta) + L'(\theta'), \\
  (\hat{\theta}_0, \hat{\theta}'_0) \triangleq \argmin_{(\theta,\theta')\in\chi_0} L(\theta) + L'(\theta').
  \end{split}
\end{equation}
Then, the generalized log-likelihood ratio can be written as
\begin{align}\label{Equ:GL}
L_{\rm{G}}(\mathcal{S},\mathcal{S}') = L(\hat{\theta}_0)+L'(\hat{\theta}'_0)-L(\hat{\theta}_1)-L'(\hat{\theta}'_1) .
\end{align}

The main difficulty in applying GLRT  is that the minimizers $(\hat{\theta}_1, \hat{\theta}'_1)$ and $(\hat{\theta}_0, \hat{\theta}'_0)$ in \eqref{Equ:GLopt} are hard to compute. In the following subsection, we propose an empirical difference test which approximates the GLRT and has reduced the computational complexity.



\subsection{Empirical Difference Test}

We need the following conditions to proceed with our analysis and establish the asymptotical normality of the MLEs \cite{van2000asymptotic}.


\begin{assump}\label{assump:MLE}
\textbf{Regularity conditions for MLE}
\begin{enumerate}
  \item   \textbf{Smoothness}: $L(\theta)$ and $L'(\theta)$ have  first, second and third derivatives for all $\theta$.
  \item \textbf{Strong Convexity}: For all $\theta$, $H(\theta)$ and $H'(\theta)$ are positive definite and invertible.
  \item \textbf{Boundedness}: For all $\theta$, the largest eigenvalues of $H(\theta)$ and $H'(\theta)$ are upper bounded by $\lambda_M$.
\end{enumerate}
\end{assump}

We note that the MLEs $(\hat{\theta}_{\rm{ML}}, \hat{\theta}'_{\rm{ML}})$ belong to either  $\chi_0$ or  $\chi_1$. If $(\hat{\theta}_{\rm{ML}}, \hat{\theta}'_{\rm{ML}})\in \chi_1$, i.e., $(\hat{\theta}_1, \hat{\theta}'_1) = (\hat{\theta}_{\rm{ML}}, \hat{\theta}'_{\rm{ML}})$,  we have $
L_{\rm{G}}(\mathcal{S},\mathcal{S}') = L(\hat{\theta}_0)-L(\hat{\theta}_{\rm{ML}})+L'(\hat{\theta}'_0)-L'(\hat{\theta}'_{\rm{ML}})> 0$. 

In addition, the worst-case false alarm probability of GLRT  is given by   $\max_{(\theta,\theta'\in \chi_0)}{\rm{P}}_{(\theta,\theta')}\{L_{\rm{G}}(\mathcal{S},\mathcal{S}') \ge \tau\}$, which we wish to upper bounded by $\alpha$. Note that $L_{\rm{G}}(\mathcal{S},\mathcal{S}')>0$ when $(\hat{\theta}_{\rm{ML}}, \hat{\theta}'_{\rm{ML}})\in \chi_1$ holds. In the following, we focus on the case where $\alpha< \max_{(\theta,\theta'\in \chi_0)}{\rm{P}}_{(\theta,\theta')}\{L_{\rm{G}}(\mathcal{S},\mathcal{S}') \ge 0\}$, i.e., a relatively small false alarm constraint $\alpha$. Thus, we just need to study the false alarm probability of GLRT when $(\hat{\theta}_{\rm{ML}}, \hat{\theta}'_{\rm{ML}})\in \chi_1$ and $\tau > 0$.

Given $(\hat{\theta}_{\rm{ML}}, \hat{\theta}'_{\rm{ML}})\in \chi_1$, it is difficult to solve for $(\hat{\theta}_0,\hat{\theta}'_0)$ in \eqref{Equ:GLopt} exactly. However, we can construct an upper bound for the GLR by approximating $(\hat{\theta}_0, \hat{\theta}'_0)$ using a linear combination of $(\hat{\theta}_{\rm{ML}}, \hat{\theta}'_{\rm{ML}})$.  Let $\Delta \hat{\theta} \triangleq \hat{\theta}'_{\rm{ML}}- \hat{\theta}_{\rm{ML}}$. Then $\|\Delta \hat{\theta}\|_2 >\rho$, 
\begin{align}\label{Equ:MLEinDiffRegions}
\widetilde{\theta}_0 = \hat{\theta}_{\rm{ML}} + \frac{\mu\Delta \hat{\theta}}{\|\Delta \hat{\theta}\|_2}, \quad
\widetilde{\theta'}_0 = \hat{\theta}_{\rm{ML}} + \frac{(\mu+\rho)\Delta \hat{\theta}}{\|\Delta \hat{\theta}\|_2},
\end{align}
where $\mu \in [0, \|\Delta \hat{\theta}\|_2-\rho]$ denotes the distance between $\widetilde{\theta}_0$ and $\hat{\theta}_{\rm{ML}}$. It can be verified that $(\widetilde{\theta}_0, \widetilde{\theta}'_0) \in \chi_0$.
Then, the GLR in \eqref{Equ:GL} can be upper bounded as
\begin{small}
\begin{align}\label{Equ:GLLRInLRM}
&L_{\rm{G}}(\mathcal{S},\mathcal{S}')
= L(\hat{\theta}_0)+L'(\hat{\theta}'_0)-L(\hat{\theta}_1)-L(\hat{\theta}'_1)\nn\\
& \le L(\widetilde{\theta}_0)+L'(\widetilde{\theta}'_0)-L(\hat{\theta}_1)-L(\hat{\theta}'_1) \nn\\
&\overset{(a)}{=} \frac{1}{2} (\hat{\theta}_1 - \widetilde{\theta}_0)^\top H(\widetilde{\theta}) (\hat{\theta}_1 - \widetilde{\theta}_0) + \frac{1}{2} (\hat{\theta}'_1 - \widetilde{\theta}'_0)^\top H'(\widetilde{\theta}') (\hat{\theta}'_1 - \widetilde{\theta}'_0)\nn\\
& = \frac{\mu^2}{2} \frac{\Delta \hat{\theta}^\top}{\|\Delta \hat{\theta}\|_2} H(\widetilde{\theta}) \frac{\Delta \hat{\theta}}{\|\Delta \hat{\theta}\|_2}\nn\\
& \qquad +\frac{(\|\Delta \hat{\theta}\|_2-(\mu+\rho))^2}{2} \frac{\Delta \hat{\theta}^\top}{\|\Delta \hat{\theta}\|_2} H'(\widetilde{\theta}') \frac{\Delta \hat{\theta}}{\|\Delta \hat{\theta}\|_2}\\
&\overset{(b)}{\le} \frac{\mu^2}{2 \sigma^2} \lambda_{\max}(H(\widetilde{\theta}) )+ \frac{(\|\Delta \hat{\theta}\|_2-(\mu+\rho))^2}{2\sigma^2}\lambda_{\max}(H'(\widetilde{\theta}')), \nn
\end{align}
\end{small}
where (a) follows from the Taylor's Theorem, $\widetilde{\theta}$ and $\widetilde{\theta}'$ denote the parameters in the corresponding remainders; and (b) follows from the fact that $H(\widetilde{\theta})$ and $H'(\widetilde{\theta}')$ are positive definite and $\frac{\Delta \hat{\theta}^\top}{\|\Delta \hat{\theta}\|_2}$ is a unit vector. Note that $\lambda_{\max}(H(\widetilde{\theta}))$ and $\lambda_{\max}(H'(\widetilde{\theta}'))$ are bounded by $\lambda_M$ in Assumption \ref{assump:MLE}.  Hence,
\begin{small}
\begin{align}\label{Equ:UpperBoundPF_LRM}
& {\rm{P}}_{\rm{F}}(\delta_{\rm{GL}}) = {\rm{P}}_{(\theta,\theta')}\{L_{\rm{G}}(\mathcal{S},\mathcal{S}') \ge \tau\}\nn \\
&\le  {\rm{P}}_{(\theta,\theta')}\Big\{\frac{\mu^2}{2 \sigma^2} \lambda_M
+ \frac{(\|\Delta \hat{\theta}\|_2-(\mu+\rho))^2}{2\sigma^2}\lambda_M\geq \tau \Big\}\nn \\
&= {\rm{P}}_{(\theta,\theta')}\big\{ \|\Delta \hat{\theta}\|_2\ge \eta \big\},
\end{align}
\end{small}
for $(\theta,\theta') \in \chi_0$. The false alarm probability of GLRT can be upper bounded by the probability that the \emph{empirical difference} $\|\Delta \hat{\theta}\|_2$ is larger than another threshold $\eta$. Note that the threshold $\eta$ can be set by letting ${\rm{P}}_{(\theta,\theta')}\big\{ \|\Delta \hat{\theta}\|_2\ge \eta \big\}\le \alpha$ for all $(\theta,\theta') \in \chi_0$, which is independent of the unknown quantities $\mu$ and $\lambda_M$.


Thus, we propose the following empirical difference test with the following structure to approximate the GLRT,
\begin{equation}\label{Equ:EDT}
{\delta}_{\rm{ED}} =
\begin{cases}
1,\quad \text{if} \,\, \|\Delta \hat{\theta}\|_2\ge \eta\\
0,\quad \text{if} \,\, \|\Delta \hat{\theta}\|_2< \eta.
\end{cases}
\end{equation}


The benefits for using ${\delta}_{\rm{ED}}$ are two-fold: 1)  Instead of constructing the more complicated GLR statistics, our EDT only requires the computation of the empirical difference $\Delta \hat{\theta}$ between the MLEs, which is more tractable in practice.  2) The distribution of the empirical difference $\Delta \hat{\theta}$ is asymptotically Gaussian, which facilitates the setting of the threshold $\eta$ to meet the false alarm constraint $\alpha$.

%


\section{Approximation for setting test threshold}
\label{Sec:threshold}

In this subsection, we provide a method based on a $\chi^2$ approximation \cite{dasgupta2008} to set the threshold $\eta$ in the EDT.

%

Since $\hat{\theta}_{\rm{ML}}$ and $\hat{\theta}'_{\rm{ML}}$ are the MLEs of $\theta$ and $\theta'$ with $n$ and $n'$ samples, respectively, we have
\begin{equation*}
\sqrt{n}(\hat{\theta}_{\rm{ML}}-\theta) \stackrel{d.}{\longrightarrow} \mathcal{N}(0,I_{{\theta}}^{-1}),\  \sqrt{n'}(\hat{\theta}'_{\rm{ML}}-\theta' )\stackrel{d.}{\longrightarrow} \mathcal{N}(0,I_{{\theta'}}^{-1}),
\end{equation*}
from the asymptotical normality of MLE \cite{van2000asymptotic}, where $I_{{\theta}}$
denotes the Fisher information matrix of the probabilistic model $p(\mathbf{z}|\theta)$.  Thus, we can approximate the distribution of $\Delta \theta$ using a Gaussian distribution $\mathcal{N}(\theta'-\theta, \Sigma_{\Delta\theta})$, where $\Sigma_{\Delta\theta} \triangleq \frac{I_{{\theta}}^{-1}}{n}+ \frac{I_{{\theta'}}^{-1}}{n'}$. In practice, $I_{{\theta}}$ and $I_{{\theta'}}$ can be estimated by replacing $\theta$ and $\theta'$ with the corresponding MLEs $\hat{\theta}_{\rm{ML}}$ and $\hat{\theta}'_{\rm{ML}}$, respectively. 


To satisfy the false alarm constraint in \eqref{Equ:NPRule}, we need to set the threshold $\eta_\alpha$ based on the following equation in the EDT,
\begin{equation}\label{Equ:OriginalThreshold}
  \max_{\theta,\theta' \in \chi_0}{\rm{P}}_{(\theta,\theta')}\{\|\Delta \theta\|^2\ge \eta_\alpha^2\} =  \alpha.
\end{equation}

The following theorem characterizes the distribution of  $\|\Delta \theta\|^2$ that results from the Gaussian approximation.
\begin{thm}
Suppose $\Delta \theta\sim \mathcal{N}(\theta'-\theta, \Sigma_{\Delta\theta})$, and the covariance matrix $\Sigma_{\Delta\theta}$ has the eigen-decomposition $\Sigma_{\Delta\theta} = P^\top \Lambda P$, where $\Lambda = {\rm{diag}}(\lambda_1,\cdots, \lambda_d)$ contains all the eigen-values, and $P$  is an orthogonal matrix.  Then,
\begin{equation}
  \|\Delta \theta\|^2 \overset{d.}{=}\sum_{i=1}^d \lambda_i(U_i+\mathbf{b}_i)^2,
\end{equation}
where $U_i \sim \mathcal{N}(0, 1)$, and $\mathbf{b} = (\sqrt{\Lambda})^{-1}(\theta'-\theta)$.
\end{thm}

The distribution of $\|\Delta \theta\|^2$ is a linear combination of independent non-central chi-squared random variables with degree of freedom of one, which does not have a simple closed form \cite{press1966linear}. We therefore propose the following approximation method to set the threshold in the EDT.
Note that
\begin{align}
  &{\rm{P}}_{\rm{F}}(\delta_{\rm{ED}}) ={\rm{P}}_{(\theta,\theta')}\Big\{\sum_{i=1}^d \lambda_i(U_i+\mathbf{b}_i)^2 \ge \eta^2\Big\}\nn\\
  & \le {\rm{P}}_{(\theta,\theta')}\Big\{\sum_{i=1}^d (U_i+\mathbf{b}_i)^2 \ge \eta^2/\lambda_{\max}(\Sigma_{\Delta\theta})\Big\},
\end{align}
for $(\theta,\theta')\in \chi_0$, and $\sum_{i=1}^d (U_i+\mathbf{b}_i)^2$ is a non-central chi-squared $\chi^2(k,\gamma)$ random variable with degrees of freedom $k=d$, and non-centrality parameter
$\gamma = \sum_{i=1}^d \mathbf{b}_i^2 \le \rho^2/\lambda_{\min}(\Sigma_{\Delta\theta})$, where the inequality follows from the fact $\|\theta'-\theta\|_2\le \rho$ under $H_0$.
Thus,
\begin{align}
  &\max_{\theta,\theta' \in \chi_0}{\rm{P}}_{(\theta,\theta')}\{\|\Delta \theta\|^2_2\ge \eta^2\} \nn \\
  &\le \max_{\theta,\theta' \in \chi_0} {\rm{P}}\Big\{\chi^2(d,\sum_{i=1}^d b_i^2) \ge \eta^2/\lambda_{\max}(\Sigma_{\Delta\theta})\Big\}.
\end{align}
We can set the threshold $\widetilde{\eta}_{\alpha}$ with the $\chi^2$ approximation \cite{dasgupta2008} using the following equation,
\begin{equation}\label{Equ:ChiSquThreshold}
  {\rm{P}}\big\{\chi^2(d,\rho^2/\lambda_{\min}(\Sigma_{\Delta\theta})) \ge \widetilde{\eta}_{\alpha}^2/\lambda_{\max}(\Sigma_{\Delta\theta})\big\}=\alpha
\end{equation}
to ensure that the false alarm probability is bounded by $\alpha$.
%

\section{NUMERICAL RESULTS}
\label{Sec:experiments}

In this section, we evaluate the performance of the proposed empirical difference test $\delta_{\rm{ED}}$  in linear regression and logistic regression models.


\noindent\textbf{Linear regression model:} The datasets $\mathcal{S}$ and $\mathcal{S}'$ are generated from the linear model $\mathbf{y} = X\theta + \mathbf{\xi}$, where $X\in \mathbb{R}^{n\times d}$ denotes the input variable,  $\mathbf{y}\in \mathbb{R}^n$ denotes the response variable and $\theta\in \mathbb{R}^d$ denotes the weight vector. We assume that all the elements in noises $\mathbf{\xi} \in \mathbb{R}^n$ are i.i.d. zero mean Gaussian random variables generated from  $\mathcal{N}(0,\sigma^2)$. Then, the Fisher information matrix  $I_{{\theta}} = XX^\top/\sigma^2$ is independent of $\theta$. In the simulations, we set the dimension $d=10$, the number of samples $n=n'=40$, $\sigma^2=1$ and $\rho=1$.

\noindent \textbf{Logistic regression model:} The datasets $\mathcal{S}$ and $\mathcal{S}'$ are generated from the following logistic model
\begin{equation}
p(y_i|\mathbf{x}_i, \theta) = \frac{1}{1+\exp(-y_i \mathbf{x}_i^\top\theta )},\ \forall (\mathbf{x}_i, y_i)\in \mathcal{S},
\end{equation}
where $\mathbf{x}_i \in \mathbb{R}^d$ denotes the feature vector, $y_i\in \{\pm1\}$ denotes the label, and $\theta \in \mathbb{R}^d, \|\theta\|_2=1$ is the normalized model parameter vector. Then, the Fisher information matrix
\begin{equation}
I_{{\theta}} = \mathbb{E}_{\mathbf{x}}\Big[\frac{1}{1+\exp(\mathbf{x}_i^\top\theta)} \frac{1}{1+\exp(-\mathbf{x}_i^\top\theta)} \mathbf{x}_i \mathbf{x}_i^\top\Big].
\end{equation}
In the simulations, we choose dimension $d=5$, the number of samples $n=n'=60$, and set $\rho$  such that the angle between $\theta$ and $\theta'$ is $\frac{\pi}{4}$.

To illustrate the performance of the proposed algorithms, we plot the probability ${\rm{P}}_{(\theta,\theta')}\{\delta=1\}$ as a function of $\|\theta-\theta'\|_2$ in all three figures, where the normalized model change $\| \theta - \theta' \|/|\rho|$ ranges from 0 to 2. Note that when $\| \theta - \theta' \|_2 < \rho$, i.e., $(\theta,\theta')\in \chi_0$, ${\rm{P}}_{(\theta,\theta')}\{\delta=1\}$ denotes the false alarm probability ${\rm{P}}_{\rm{F}}(\delta)$ (in the left side of the figures). In contrast, when $\| \theta - \theta' \|_2 > \rho$, $(\theta,\theta')\in \chi_1$ and ${\rm{P}}_{(\theta,\theta')}\{\delta=1\}$ denotes the detection probability ${\rm{P}}_{\rm{D}}(\delta)$ (in the right side of the figures). Thus, the plot of ${\rm{P}}_{(\theta,\theta')}\{\delta=1\}$  provides us with an illustration of the test performance under both hypotheses with different model parameters.

\begin{figure}
\begin{minipage}[t]{0.5\textwidth}
  \centering
  \includegraphics[width=7.9cm]{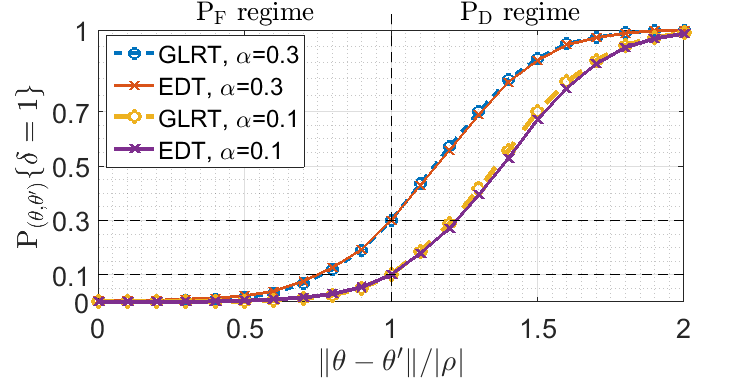}\\
    \vspace{-0.3cm}
  \caption{Comparison of the performances of the GLRT and EDT for the linear regression model.}\label{Fig:linear_test_str}
\end{minipage}
\\
\begin{minipage}[t]{0.5\textwidth}
  \centering
  \includegraphics[width=7.9cm]{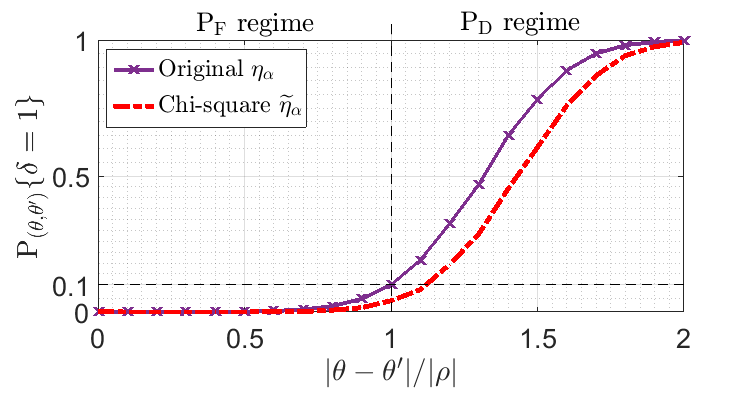}\\
  \vspace{-0.3cm}
  \caption{Comparison of the performance of EDT with the threshold $\eta_\alpha$ and the $\chi^2$ approximation $\widetilde{\eta}_\alpha$, for the linear regression model with $\alpha=0.1$.}\label{Fig:linear_approx}
\end{minipage}
\\
\begin{minipage}[t]{0.5\textwidth}
  \centering
  \includegraphics[width=7.9cm]{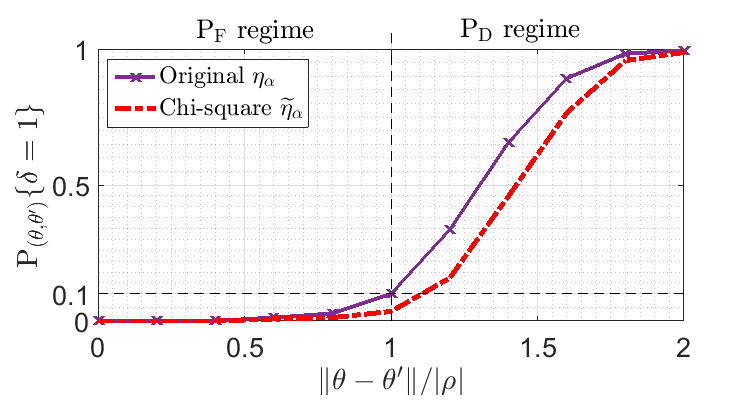}\\
  \vspace{-0.3cm}
  \caption{Comparison of the performance of EDT with the threshold $\eta_\alpha$ and the $\chi^2$ approximation $\widetilde{\eta}_\alpha$, for the logistic regression model with $\alpha=0.1$. }\label{Fig:logistic_approx}
\end{minipage}
\end{figure}

To verify the approximation of the GLRT with the proposed EDT, we first compare the performance of these tests for the linear regression model (the GLRT is not computationally feasible for logistic regression) for two values of the false alarm constraint $\alpha=0.1$ and $\alpha=0.3$. The thresholds of these tests $\eta_\alpha$ are set using 1000 runs of Monte-Carlo simulations such that the false alarm probabilities are equal to $\alpha$ as in \eqref{Equ:OriginalThreshold}. It is shown in Fig. \ref{Fig:linear_test_str} that the difference between the performance of EDT and that of GLRT is negligible with only $n=n'=40$ samples, which justifies the use of EDT. We note that when $\| \theta - \theta' \|/|\rho| = 1$, it is impossible to distinguish $H_0$ and $H_1$ even if the number of samples $n$ and $n'$ go to infinity, i.e., the probabilities of false alarm and detection are both equal to $\alpha$ in this case.

%




Fig. \ref{Fig:linear_approx} and Fig. \ref{Fig:logistic_approx} compare the performance of EDT with the threshold $\eta_\alpha$ computed by 1000 runs of Monte-Carlo simulations in \eqref{Equ:OriginalThreshold}, and the threshold $\widetilde{\eta}_\alpha$ set by the proposed $\chi^2$ approximation in \eqref{Equ:ChiSquThreshold}, respectively, when $\alpha=0.1$. It can be observed that in both linear regression and logistic regression cases, the non-central chi-squared approximation in \eqref{Equ:ChiSquThreshold} provides conservative estimates of the test thresholds $\eta$, thereby ensuring that the false alarm constraint is met. 

\clearpage

\bibliography{model_change}
\bibliographystyle{IEEEbib}

\end{document}